\documentclass[conference]{IEEEtran}
\usepackage{times}

\usepackage[numbers]{natbib}
\usepackage{multicol}
\usepackage[bookmarks=true]{hyperref}

\usepackage{amsmath,amssymb,amsfonts}
\usepackage{algorithmic}
\usepackage{graphicx}
\usepackage{textcomp}
\usepackage{xcolor}

\usepackage{soul}
\usepackage{caption}
\usepackage{subcaption}
\usepackage{dialogue}
\usepackage{comment}
\usepackage{balance}

\usepackage[T1]{fontenc} 
\usepackage{setspace} 
\usepackage{dialogue}
\usepackage{calc} 

\newlength{\widestname}
\makeatletter
\renewenvironment{dialogue} {%
    \begin{list}{} {%
        \setlength\itemsep{\z@ \@plus .5ex} %
        \setlength{\parsep}{\parskip} %
        \setlength{\rightmargin}{0pt} 
        \setlength{\labelwidth}{\widestname} 
        \setlength{\labelsep}{0.5em} 
        \setlength{\leftmargin}{\labelwidth} 
        \addtolength{\leftmargin}{\labelsep} 
        \defcommand\speak [1] {\item[{##1}]} 
        
      }%
      \PreDialogue\relax
    }{%
  \end{list}%
  }
\makeatother

\begin{document}

\title{A Robot-Assisted Approach to Small Talk Training for Adults with ASD}

\author{
Rebecca Ramnauth$^{1}$, Dra\v{z}en Br\v{s}\v{c}i\'{c}$^{2}$, Brian Scassellati$^{1}$\\
$^{1}$Department of Computer Science, Yale University, USA\\
$^{2}$Graduate School of Informatics, Kyoto University, Japan\\
{\tt\small rebecca.ramnauth@yale.edu}
}

\maketitle

\begin{abstract}
From dating to job interviews, making new friends or simply chatting with the cashier at checkout, engaging in small talk is a vital, everyday social skill. For adults with Autism Spectrum Disorder (ASD), small talk can be particularly challenging, yet it is essential for social integration, building relationships, and accessing professional opportunities. In this study, we present our development and evaluation of an in-home autonomous robot system that allows users to practice small talk. Results from the week-long study show that adults with ASD enjoyed the training, made notable progress in initiating conversations and improving eye contact, and viewed the system as a valuable tool for enhancing their conversational skills.
\end{abstract}
\begin{IEEEkeywords}
human-robot interaction, autism spectrum disorder, socially assistive robotics, social-skills training
\end{IEEEkeywords}

\IEEEpeerreviewmaketitle

\section{Introduction}

Imagine a scene where three coworkers are engaging in small talk at the beginning of their workday. One of them is Alex, who has Autism Spectrum Disorder (ASD), a neurodevelopmental condition that often makes it challenging to understand and interpret social cues \cite{american2013diagnostic}.

\begin{dialogue}
    \speak{C} Hey everyone, how's it going?
    \speak{B} Hi! Not bad... just trying to power through this Monday. How about you, Alex?
    \speak{A} Monday is okay.
    \speak{C} Good to hear. Anything exciting happen this weekend?
    \speak{B} Yeah, I finally tried that new restaurant. It was fantas---
    \speak{A} I watched a movie.
    \speak{C} Glad you liked the restaurant, Ben. It's my kids' favorite spot these days... What movie did you watch, Alex?
    \speak{A} ``The Martian.''
    \speak{B} I love that one! Matt Damon is awesome.
    \speak{A} \textit{No response.}
\end{dialogue}

At first glance, this brief example of a typical interaction appears unremarkable. It represents the everyday small talk that occurs regularly in many workplaces. For workers with ASD, however, such apparently ``easy'' interactions may present a real challenge. In this example, while Alex responds to direct questions, the responses are brief and lack elaboration. Alex's responses provide minimal information rather than actively participating in the flow of the conversation. Additionally, Alex's lack of response to the last prompt may suggest difficulty in extending or sustaining the dialogue.

Although workers with ASD are often highly trained and skilled in job-specific tasks, they frequently face challenges with social interactions in the workplace. M\"{u}ller et al. \cite{muller2003meeting} and Hurlbutt \& Chalmers \cite{hurlbutt2004employment} found that many of their participants with ASD, despite completing graduate-level coursework, were employed in positions for which they were over-qualified, such as food services or low-level administrative roles. This underemployment is further highlighted in the National Longitudinal Transition Study \cite{roux2013postsecondary}.


However, the ability to perform work tasks is only the tip of the iceberg when it comes to workplace success. Interpersonal skills are proven to be more significant predictors of overall success \cite{holmes2005small, joseph2021teaching, ohl2017predictors}. Hurlbutt \& Chalmers \cite{hurlbutt2004employment} found that workers with ASD often attributed job-related challenges to social factors rather than the work-specific tasks themselves. In their study, one interviewee noted, ``Jobs usually are 80\% social (conversation, lunch, breaks, chit-chat) and 20\% work.'' Furthermore, adults with ASD have reported that difficulty engaging in ``social niceties,'' such as small talk, led to feelings of isolation and alienation in the workplace \cite{hurlbutt2004employment, muller2003meeting}.

Unlike the functional aspects of other on-the-job communication, which may focus on conveying information or assistance, small talk is considered purely social. It acts as a social lubricant, fostering rapport, mutual understanding, and trust \cite{coupland2014small}, and is widely recognized as a key facilitator in building and maintaining relationships. In professional settings, small talk is considered an essential tool for networking success and establishing positive first impressions \cite{pullin2010small, bristoll2015small}. It is even regarded as a vital skill that should be targeted in communication therapy for various populations \cite{holmes2005small, joseph2021teaching, garrels2019getting}.

To better support adults with ASD, it is important to develop accessible and targeted opportunities for improving interpersonal skills, such as small talk. Social robotics has the potential to enhance existing training initiatives by improving access to personalized, socially-situated, and physically co-present interactions \cite{mataric2016socially, scassellati2012robots, cabibihan2013robots, ramnauth2022social}. Physically present robots have proven effective in improving users' social abilities \cite{scassellati2018improving, ramnauth2022social}, providing benefits such as enhanced engagement, improved social confidence, and greater motivation to participate---outcomes that are notably more pronounced than those achieved with non-embodied technologies \cite{nadel2022autism}. 

Furthermore, research has established that robots for ASD interventions can result in positive and productive outcomes \cite{scassellati2012robots}. Social robots have demonstrated general effectiveness in enhancing verbal communication skills \cite{kozima2005interactive, robins2009isolation, kanero2018social}, including the ability to engage in everyday dialogue \cite{babel2021small, ise2021social, nichols2022hey}. Additionally, participants with ASD in recent studies have described such robot-assisted training as relevant and useful in their workplace experiences \cite{mckenna2020sorry, ramnauth2022social}.

Thus, leveraging our prior successes in developing robots for ASD interventions \cite{scassellati2012robots, scassellati2018improving, ramnauth2022social}, we developed an autonomous training system that helps adults with ASD improve their small talk skills. Given the literature on the communicative difficulties of ASD (Sec. \ref{sec:background}), we begin by examining the extent to which small talk is considered a desired social skill (Sec. \ref{sec:survey}). These insights inform our design requirements (Sec. \ref{sec:goals}) and guide the development of our prototype (Sec. \ref{sec:prototype}). In that formative study, we investigate initial impressions, perceptions of robot-assisted training, and anticipated outcomes for users with ASD. We then present findings from a week-long, in-home deployment (Sec. \ref{sec:deployment}), highlighting how users with ASD received and engaged with the robot-assisted training.

\section{Related Work} \label{sec:background}

Atypical communicative behaviors are key diagnostic criteria for ASD \cite{american2013diagnostic}, often presenting as limited eye contact, difficulty understanding sarcasm or abstract language \cite{pexman2019addressing}, and challenges in grasping the social rules that govern everyday interactions \cite{silver2022perspectives}. However, everyday, casual conversations are a pervasive aspect of daily life, whether it is chatting with a neighbor about the weather, maintaining friendships, or making a positive impression on the first day of a new job.

Unlike more structured interactions, which can be formalized to teach more easily as a script \cite{bellini2008social, akers2016synthesis}, small talk demand quick thinking, social flexibility, and the ability to interpret subtle cues such as tone, timing, and context. Training to improve such skills in adulthood presents unique challenges compared to childhood, as many social habits and patterns are already established by the time individuals reach adulthood \cite{dematteo2012social}. These patterns may manifest in the development of strategies to avoid social situations entirely \cite{bejerot2014social}; hence, the skill development that typically occurs through ongoing social interaction may not have been fully realized or practiced. Furthermore, while children have more opportunities for structured social learning and development through school or therapy, adults with ASD may have fewer chances to actively develop these skills \cite{perkins2012into, gerhardt2011addressing, graetz2010autism}.

As a result, interventions for adult learners often require more personalized approaches, targeting specific barriers to communication and focusing on building confidence in real-world conversational contexts. Therefore, in this section, we overview the structure and value of small talk, outline formal intervention strategies that may inform the pedagogical design of small talk training, and highlight the potential for robot-assisted social skills training for ASD. This summary of the literature ultimately reveals a gap: while there is a recognized need for small talk skills, there is limited input from adults with ASD themselves regarding the challenges they face in adulthood and few opportunities for targeted training. To address this gap, we conducted a survey to gather insights from adults with ASD (Sec. \ref{sec:survey}).

\subsection{Structure and Value of Small Talk}
While the boundaries of conversation types are fluid, ``small talk'' has a recognized currency in sociolinguistics and communication studies \cite{coupland2014small}. It refers to informal, light-hearted exchanges focused on building social connections rather than conveying substantial detail, often covering general, non-controversial topics like the weather and hobbies. 

Small talk does not have a strict formula, as it is inherently flexible and context-dependent. However, a typical small talk dialogue follows the general sequence of conversation, beginning with a greeting and ending with a closing remark, while emphasizing specific characteristics at each stage \cite{jaworski2014silence, coupland2014small, ramnauth2024grounded}:

\begin{enumerate}
    \item \textbf{Greetings and openers}: Initiating the conversation with a greeting 
    or commenting on a shared experience such as the weather or the immediate environment.
    \item \textbf{General topics}: Discussing non-controversial and general topics such as hobbies, interests, or recent events.
    \item \textbf{Reciprocity}: Both participants take turns sharing and responding, maintaining a balanced, equitable, and relevant participation in the conversation.
    \item \textbf{Closure}: The conversation ends with a closing remark, such as indicating appreciation or a future interaction.
\end{enumerate}

For any individual, the characteristics of small talk highlight the subtlety and skill needed to navigate this form of conversation effectively. For adults with ASD, there is a notable overlap between the challenges inherent in small talk and the broader difficulties they report facing in everyday social interactions. We discuss this overlap further in Sec. \ref{sec:survey}.


\subsection{Current Approaches to Small Talk Training}
Addressing the unique challenges of small talk for individuals with ASD requires targeted interventions. Although there are currently no programs focused solely on small talk, many broader training initiatives include elements that indirectly support small talk competency. Moreover, while the limited availability of such training for adults with ASD has not been widely explored or critiqued in the literature, the following sections review broader communication programs, which are primarily focused on \emph{children} with ASD. We then examine how these established methods could potentially address the specific challenges of small talk for adults with ASD.

\subsubsection{Didactic Approaches}
Didactic approaches, or classical Applied Behavior Analysis (ABA), break skills into smaller components and train each through highly structured, drill-like practice \cite{paul2008interventions, duggal2020works, dematteo2012social}. While didactic methods have proven effective in various intervention studies for ASD \cite{wermer2018efficacy, dehqonova2022importance}, they heavily depend on teacher guidance, prompted responses, and contrived reinforcement methods \cite{paul2008interventions}. An inherent limitation of didactic methods lies in their tendency to encourage passive communication, wherein individuals respond to prompts but may struggle to initiate communication or apply learned behaviors beyond the specific training context.

\subsubsection{Naturalistic Approaches}
Contemporary or naturalistic ABA strives to incorporate interventions into an individual's everyday environment. While these approaches retain some level of teacher direction, focusing on predetermined goals, they emphasize intrinsic reinforcements, such as personal motivation or social reinforcement. Studies directly comparing didactic and naturalistic approaches have indicated certain advantages of the latter, including better retention and broader application of newly acquired communication skills \cite{delprato2001comparisons, paul2008interventions}. \emph{Milieu teaching}, a subset of naturalistic ABA, integrates training into everyday environments to effectively promote spontaneous communication and initiation in individuals with ASD \cite{christensen2013impact, mancil2009milieu, kaiser2000effects, ogletree2020selective}. This method encourages skill development through activities that occur organically throughout the day, rather than being confined to a designated ``therapy time.'' 

\subsection{Robot-Assisted Social Skills Training for ASD}
There is considerable evidence that technology-driven, practice-based interactions can enhance social skills in adults with ASD \cite{burke2010social, ramnauth2022social}. However, robots offer distinct advantages over other technologies by providing a physical, embodied presence that naturally demands a social response \cite{bainbridge2008effect, leyzberg2011robots}. A socially assistive robot (SAR) may feature human-like attributes, such as a face capable of mimicking human expressions or the ability to make eye contact, both of which can elicit social responses from users. This presence allows users to engage in consistent, real-world practice, providing access to repeatable, co-present social interactions that may be challenging to replicate with human therapists \cite{scassellati2012robots} or in naturally-occurring, everyday situations. 

Several studies have demonstrated the effectiveness of robot-assisted therapy for ASD, with evidence showing improvements in a variety of social and behavioral outcomes. For instance, research has indicated that SARs can foster prosocial behaviors \cite{diehl2012clinical}, sustain attention \cite{ramnauth2025gaze}, elicit spontaneous and appropriate social behavior \cite{scassellati2018improving}, reduce stereotyped and repetitive behaviors \cite{srinivasan2015comparison}, optimize cognitive learning gains \cite{robins2012embodiment}, and heighten social engagement \cite{scassellati2012robots, pennisi2016autism}. 

While the majority of these studies have focused on children, the growing body of research supports the efficacy of SARs for adults with ASD as well \cite{mckenna2020sorry, ramnauth2022social}. A robot designed for social skills training could therefore be a valuable tool for adults with ASD, offering a safe, consistent, and adaptive platform for practicing and refining their skills. 

\section{Survey on the Need for Small Talk Training} \label{sec:survey}
To supplement insights from existing literature, we conducted an initial survey to explore the firsthand experiences and challenges faced by adults with ASD. This needs-finding survey aimed to identify specific conversational areas where interventions are not only desired but also deemed most valuable by this population. We also contextualize the survey results within existing research on ASD. 

Fifty adults with ASD (22 men and 28 women)\footnote{This sample's nearly balanced gender ratio differs from the typical 3:1 male-to-female ratio in ASD diagnoses \cite{loomes2017male}. This discrepancy may be due to factors such as increased awareness of underdiagnosis in women, recent increases in diagnosis rates, or specific recruitment material or methods in Prolific, which likely attracted a more diverse group of participants.}, ranging from 20 to 55 years ($M = 31.4$, $SD = 10.1$) responded to our online survey administered on Prolific \cite{palan2018prolific}. All participants reported having been clinically diagnosed with ASD, either in childhood ($N = 18$) or adulthood ($N = 32$). The survey objectives, design, and analysis were preregistered \cite{ramnauth_brscic_scassellati_2024}. 


\subsection{Small Talk Skills \& ASD}\label{sec:small-talk-and-asd}
While the survey was designed to broadly explore conversational skills in the context of adult social interactions, the majority of respondents emphasized skills closely associated with small talk. When asked in an optional, open-ended prompt about which conversational skills they wished to improve, 48 adults with ASD ($96\%$) specifically expressed a desire to enhance skills that are central to small talk interactions. These reported challenges and desired skills are grouped into five themes as outlined below.

\textbf{Difficulty Initiating Conversations.}\label{sec:initiating}
Individuals with ASD may find it challenging to initiate conversations, particularly in unfamiliar social situations \cite{paul2009conversational}. Initiating conversations often relies on subtle social cues, such as recognizing when to engage, gauging the appropriate timing for greetings, and responding in a manner that aligns with the context. Among the adults with ASD surveyed, 27 individuals ($54\%$) reported initiating conversations is a skill they wished to improve. 

\textbf{Limited Interest in Non-Specific Topics.}\label{sec:topic}
Individuals with ASD often exhibit a preference for structured and predictable interactions \cite{petrolini2023does, paul2009conversational}. The open-ended and non-specific nature of small talk may be discomforting for individuals who prefer environments with well-defined rules and expectations.

Furthermore, individuals with ASD may exhibit a strong interest in specific topics or subjects \cite{nadig2010does, schwartz2009conversational}, often preferring in-depth discussions over broad, superficial exchanges. $P_{19}$ stated, ``I'm not able to come up with and react to lighthearted banter quickly enough, so I come across as quiet and serious. If I don't know someone well enough, I have no idea what kind of information they would like to receive [...] so I just stay quiet'' The preference for depth and detail can make it challenging to engage in the more superficial content typical of small talk. Among those surveyed, 20 individuals ($40\%$) highlighted the ability to discuss general topics beyond their own interests as a specific skill they wished to improve.

\textbf{Lack of Reciprocal Exchange.}\label{sec:maintaining}
Small talk relies on a timely back-and-forth exchange of information and active listening. Individuals with ASD often face difficulties in maintaining reciprocity during conversations, leading to challenges in sustaining the flow of dialogue \cite{nadig2010does, ochi2019quantification, van2012measuring, koegel2014using}. Among the adults with ASD surveyed, 39 ($78\%$) expressed a desire to improve their ability to maintain balanced and reciprocal conversations.

\textbf{Insistence on Topic Sameness.}\label{sec:transitioning}
Transitioning smoothly between different topics or concluding the conversation is a skill often required in small talk. Adults with ASD may find it challenging to navigate these transitions, leading to potential disruptions in the flow of conversation \cite{ying2018systematic, geurts2009paradox}. Among the adults with ASD surveyed, 18 individuals ($36\%$) reported that smoothly transitioning between different topics is a valuable conversational skill that they wished to improve.

\textbf{Social Anxiety.}
The nuances of social cues can exacerbate feelings of anxiety, leading to the avoidance of social situations altogether \cite{balderaz2020social}. Although reporting wanting to improve in the aforementioned skills, $P_{27}$ shared, ''I feel a debilitating consciousness about my eye contact and posture [...] Even if I end up talking, I'm never sure whether it was the right thing.'' Small talk requires maintaining conversational flow while interpreting nonverbal cues, which can heighten anxiety. On a 7-point Likert scale (1 = highly uncomfortable, 7 = highly comfortable), adults with ASD reported feeling moderately uncomfortable ($M = 2.3$, $SD = 1.4$). 35 participants ($70\%$) identified managing anxiety as a skill they wished to improve.

\subsection{Methods \& Challenges to Improvement}
Improving small talk skills is seen as highly valuable by adults with ASD, as it plays a critical role in fostering social connections and achieving personal and professional goals. In an optional open-ended response, $40$ of the $50$ surveyed adults with ASD described the value of this kind of informal conversation in specific aspects of their own lives. We conducted a thematic analysis on their responses, ultimately grouping the descriptions into three primary themes: making new friends ($60\%$), dating ($33\%$), or finding and maintaining a job ($60\%$). Adults with ASD explained that conversations such as ``small talk'' would be useful to ``getting to know people and keeping them interested'' ($P_{16}$), and in ``job interviews as [...] you have to use a specific but casual enough talking structure to be considered adequate'' ($P_{25}$) or ``socially competent'' ($P_{48}$). 

Yet, few adults with ASD (\emph{N} = 10, $20\%$) reported undergoing formal training in casual conversations through a vocational program, an online class, or coaching from a therapist specialized in interpersonal communication. Despite actively seeking training, the majority of adults ($78\%$) reported having limited or no formal opportunities ($N = 10$) or relying on informal methods ($N = 29$), such as improving their conversational skills through observation or seeking feedback from friends and family. 

Generally, humans learn to analyze the interactions they observe and deduce the rules from everyday, naturally-occurring exposure. In free-form responses on the methods that were most beneficial for conversational skill development, 32 adults with ASD ($64\%$) described learning through mere practice and observation beyond or absent of any explicit training. However, natural exposure to an adequate range of real-world interactions, such as the kinds of conversations typical of a workplace, is unlikely for some adults with ASD. To this, $P_{3}$ explained that improving in the small talk skills would be valuable for ``making friends and going to classes'', but ``part of the reason I don't do this much is because I don't put myself in situations where I would meet people.''

In summary, small talk presents significant challenges for individuals with ASD, particularly in initiating, maintaining, and transitioning within casual conversations. As one participant, $P_{37}$, explained, ``Doing it [small talk] \emph{is} the challenge. Knowing when you should make conversation, knowing the unsaid cues of if a conversation should continue or end. Knowing what to talk about, not saying too much when asked a question, making sure you ask the next question after. It is a very mentally draining process where I have to evaluate many different factors.'' In fact, nearly all participants ($98\%$) in this preliminary survey expressed a desire to improve specific skills closely related to small talk. Understanding these challenges is essential for developing and improving access to effective interventions tailored to the unique needs of adults with ASD.

\section{Formative Study}\label{sec:prototype}

A robot for small talk training should possess proficient small talk ability: balancing conversational succinctness and depth, maintaining an expressive yet appropriate tone, and generating relevant and open-ended responses. Prior research has explored the feasibility of developing robots for small talk interactions \cite{ramnauth2024grounded}. While large language models (LLMs) show substantial potential for enabling natural language capabilities in robots, achieving seamless and contextually appropriate casual dialogue for repeated interactions remains a challenge. Therefore, the first step in developing a robot platform for small talk training is to address this challenge and create a system capable of both generating and evaluating small talk. 

\begin{figure}[t]
    \centerline{\includegraphics[width=\linewidth]{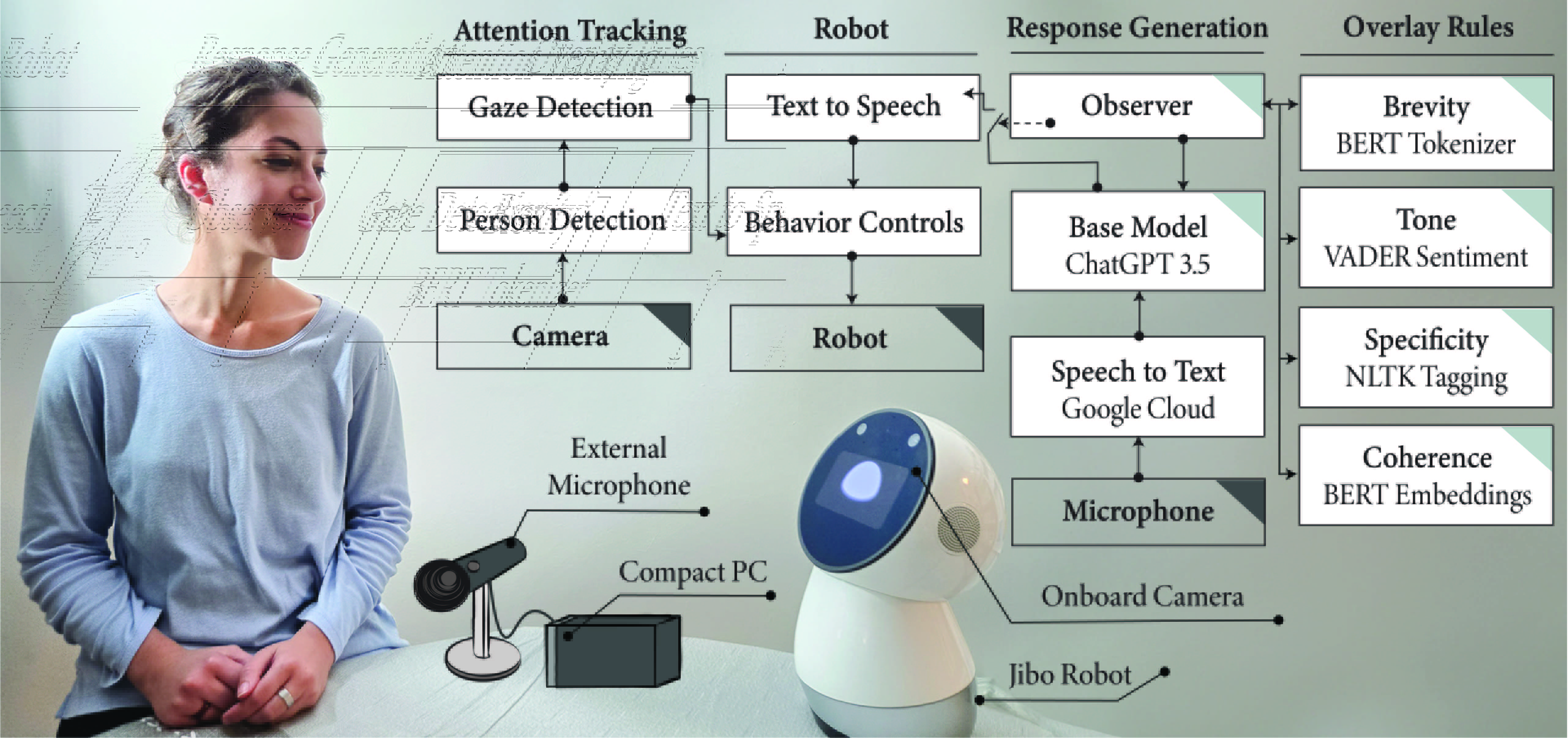}}
    \caption{\textbf{Grounded Observer}. Following the framework outlined by Ramnauth et al. \cite{ramnauth2024grounded}, we implemented an observer model that enables LLMs to sustain authentic small talk interactions effectively. We then applied this model to a robot, creating a prototype of the training system, and surveyed adults with ASD to gather feedback on whether they would use such a system to improve their small talk skills.}
    \label{fig:prototype}
    \vspace{-12pt}
\end{figure}

Good conversation is believed to arise from the control of low-level attributes \cite{wei2022emergent, ramnauth2024grounded}. We implemented a grounded observer model---an LLM instance that ``observes'' ongoing conversations to evaluate whether responses from the ``speaking'' model adhere to the quantifiable small talk criteria described in Ramnauth et al. (i.e., brevity, tone, coherence, and topic non-specificity) \cite{ramnauth2024grounded}. If the generated response aligns with these criteria, it is relayed; otherwise, the observer generates a revised system prompt and returns it to the speaking model as feedback. This feedback redirection allows the system to self-correct when drifts in conversational behavior are detected. Our implementation is depicted in Fig. \ref{fig:prototype}.

Prior work has not only showed the inadequacy of an ``out-of-the-box'' LLM for sustaining small talk, but also the observer's robustness in  real-time human-robot interactions \cite{ramnauth2024grounded}. Ramnauth et al. \cite{ramnauth2024grounded} found that users were dissatisfied with the baseline LLM's responses, noting its overly assistive and verbose tendencies, which led to conversations described as ``rambling'' and ``like speaking to a wall.'' In contrast, the observer-enabled system produced conversations that users described as ``natural'' and ``like a casual chat with someone you haven't met before or haven't seen in a long time.'' 


In this present study, we evaluated the robot prototype with the 50 adults with ASD who participated in our needs-finding survey (Sec. \ref{sec:survey}). The goal of presenting this early prototype was to assess whether the observer-enabled small talk reflected relevant real-world conversations and to determine whether adults with ASD would be receptive to both the robot and training. Three randomly selected video excerpts of user interactions were shown. 
In open-ended feedback, $84\%$ of participants (\emph{N} = 42) reiterated the value of practice-based small talk training for improving confidence ($P_{37}$), building a habit of engaging in interactions ($P_{3}$), and exercising strategies for handling dynamic situations ($P_{24}$). Participants also noted that having the robot at home allows for ``practice in a safe environment'' ($P_{14}$) with a ``non-judgmental partner'' ($P_{33}$). 

Yet, some participants (\emph{N} = 12) expressed wariness to having the robot in their homes. 
$P_{31}$ noted that, while the robot offers non-judgmental practice, ``other people may overhear my conversations.'' In light of these concerns, several design criteria were established. For example, we aim to design a portable system that operates within a preferred time window, thus allowing users to control when and where interactions take place. Additionally, the system should initiate training only when the user is not engaged in other social activities. The detectors that enable this functionality are described in Sec. \ref{sec:training}. This design must prioritize user comfort and security, ensuring a personalized and respectful experience. We expand on our design considerations in the following section. 

\section{Design Goals for Robot-Assisted Training} \label{sec:goals}

Addressing the specific challenges that adults with ASD face calls for targeted small talk interventions. This section lists the core design principles for our robot-assisted training.

\subsection{System Design Objectives}
Adhering to the tenets of milieu teaching (Sec. \ref{sec:background}-C), we designed a robot for small talk training with these objectives:

\textbf{In a Natural Setting}. The system should be tailored for in-home training. This enables users to interact without concern for potential stigma from colleagues. It also eliminates the need for approval or disclosure of a diagnosis to others.

\textbf{Realistic Interactions}. The robot should offer timely and responsive reactions to the user's communication attempts. The robot system must deliver authentic interactions that mirror the appropriateness and style of real-world small talk, responding in real-time and expressing human-like behaviors, including naturalistic gaze, movement, and speech. Additionally, training sessions should not be confined to a designated, scheduled ``therapy time,'' but should occur more organically throughout a portion of the user's day. 

\textbf{User-Led Interactions}. The robot's design should empower users to take the lead in interactions by responding to their cues and adjusting its behavior accordingly. A social robot inherently facilitates this objective through its physical presence in the training experience, making it challenging for users to ignore even non-verbal prompts for interaction \cite{bainbridge2008effect, fiore2013toward, ramnauth2022social}.


\textbf{Autonomous Behavior}. Training must be entirely autonomous, eliminating the need for technical expertise to adjust or control the system once it is given to the user. Although similar systems for ASD have been designed for clinical or laboratory settings where environmental conditions can be controlled or planned for \cite{scassellati2018improving}, the home is a dynamic, unstructured environment that demands more complex sensing.

\subsection{Training Design Objectives}
To design a training method, we break down small talk interactions into smaller, more manageable components and tailor the training method to the unique needs of individuals with ASD. Given the challenges of small talk for individuals with ASD (Sec. \ref{sec:initiating}), the training focuses on four primary components: initiating a conversation, discussing a non-specific topic, maintaining a flow of dialogue, and appropriately transitioning to a new topic or concluding the interaction. 

\begin{figure*}[t]
   \centerline{\includegraphics[width=\textwidth]{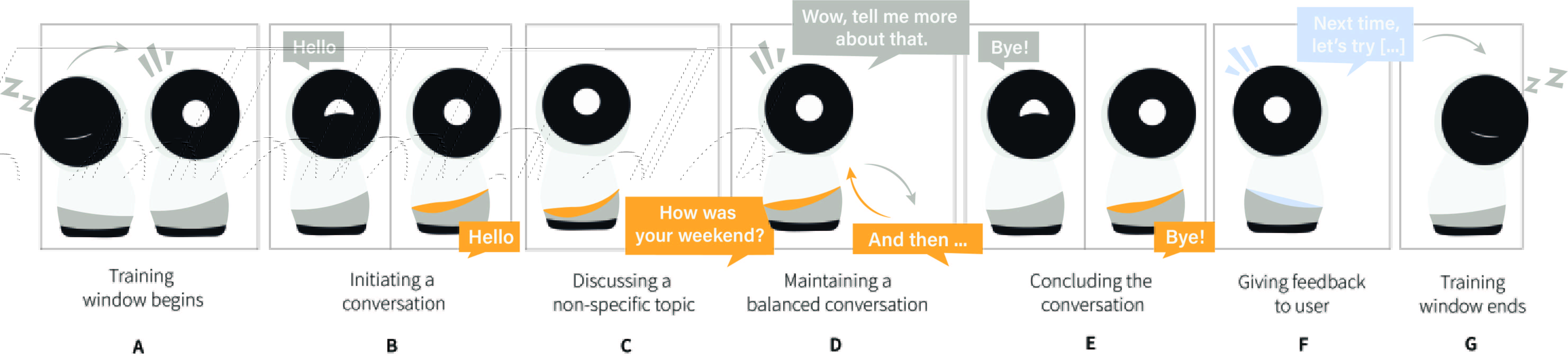}}
    \caption{\textbf{Training Sequence}: A training session unfolds in distinct stages: (A) The robot ``wakes up'' to signal the start of the session, (B) encouraging users to initiate with a greeting. Users practice small talk skills, including (C) discussing non-specific topics, (D) maintaining balanced dialogues, and appropriate transitions (E). Users complete several conversations, with the robot giving micro-level feedback after each conversation (B-F). At the end of session, the robot gives the user macro-level, overall feedback (F). Outside of the specified training window, the robot performs an idling behavior (G).}
    \label{fig:sequence}
    \vspace{-12pt}
\end{figure*}

A tiered training method that models the natural flow of dialogue would sequentially address each component, starting with a simple greeting. The robot responds promptly and contingently, reinforcing positive behaviors. Progressing through the tiers, users tackle more complex aspects, such as introducing non-specific topics or maintaining conversational flow. This approach, illustrated in Fig. \ref{fig:sequence}, encourages users to practice the core components of small talk within each training interaction. Additional considerations include:

\textbf{Fading Robot-Initiated Prompts}. The robot ``wakes up'' from its idling state to draw attention to the beginning of a training session (Fig. \ref{fig:sequence}-A). In initial interactions with the user, the robot provides an explicit verbal prompt, such as saying, ``The training window has started. Remember to make eye contact and greet me.'' As the user becomes accustomed to the initiation process, the robot gradually minimizes the verbal prompt: ``The training window has started.'' with a less pronounced emphasis. Over time, it is expected that the user initiates to interaction without any specific cue (Fig. \ref{fig:sequence}-B).



\textbf{Relevant Feedback}. The robot should offer clear, supportive feedback, highlighting both user successes and areas for improvement. This encourages users to reflect on their conversations and identify ways to enhance their skills.


\begin{figure}[t]
    \centerline{\includegraphics[width=0.75\linewidth]{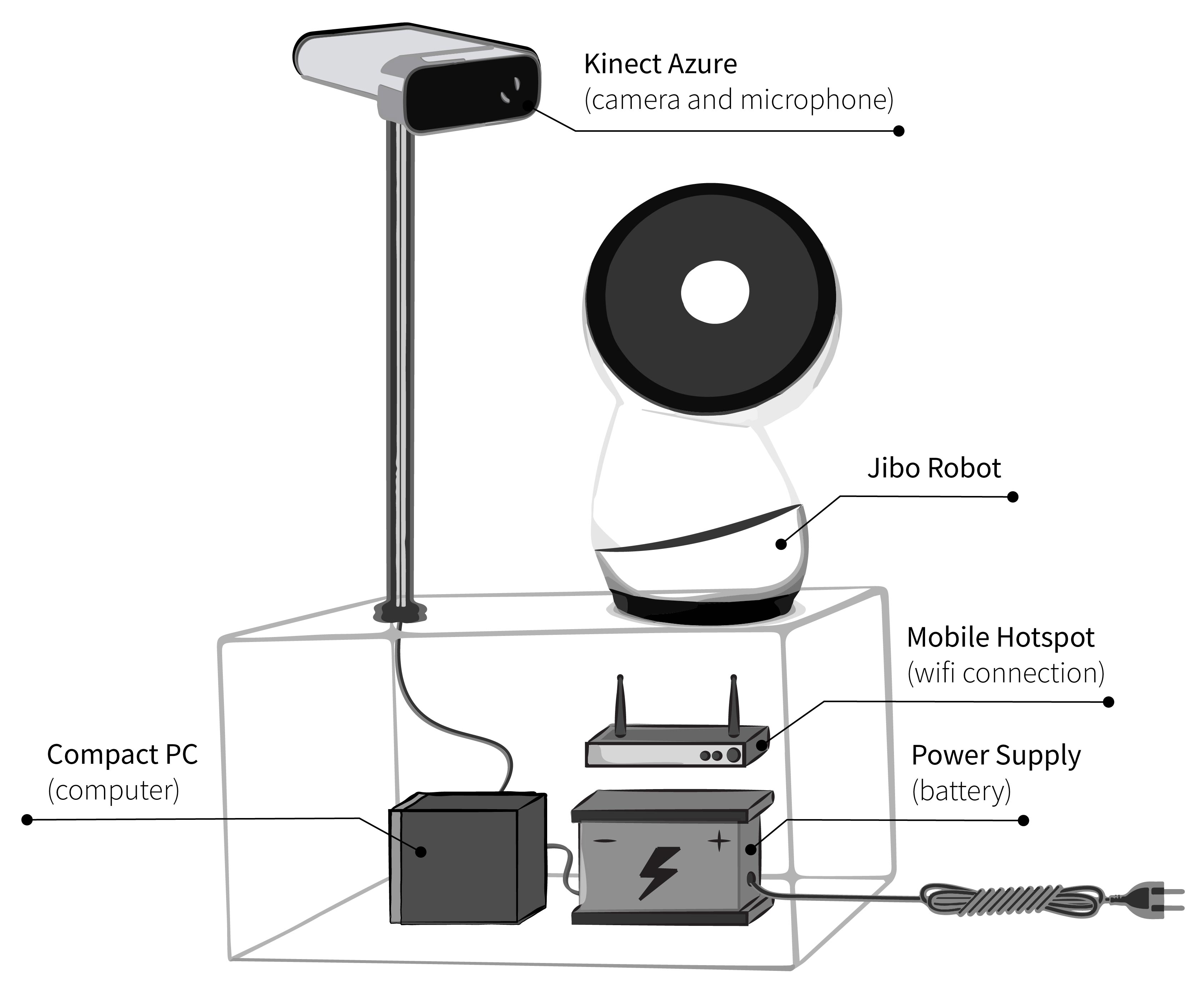}}
    \caption{\textbf{System Hardware}. The system has a battery, compact computer, and hotspot that are contained in a hard plastic case. An external camera and microphone are mounted on a mast.}
    \label{fig:hardware}
    \vspace{-12pt}
\end{figure}

\textbf{Session Length}. The optimal amount of training will vary among individuals, but a general guideline is to engage in daily practice. Consistent, short sessions with the robot would be more beneficial than long, less frequent sessions \cite{schneider1983attention, zimmerman2006development}. 

\section{System}
This section overviews the hardware and software components that address our design goals (Sec. \ref{sec:goals}) for a fully autonomous, robot-assisted training system for small talk. 

\subsection{Hardware}

Our system consisted of five primary hardware components, as shown in Fig. \ref{fig:hardware}. We used the Jibo robot \cite{Jibo}, which stands 11 inches tall and features three full-revolute axes for 360-degree movement. Jibo's onboard capabilities enabled naturalistic gaze and movement. A compact PC communicated with other hardware, monitored the system, and served as local data storage during deployment. An Azure Kinect camera was mounted 2 inches above Jibo's head to maximize the visual field and audio capture during training.

The system was designed for self-reliance, equipped with a mobile router that provides a prepaid internet connection for continuous WiFi and automatic cloud-based data synchronization. An uninterruptible power supply served as the charging station. With these components, the setup was plug-and-play, requiring only the charging station to be plugged in. To enhance ergonomic and accessible design, non-interfaceable components were encased in the container on which the robot was placed, reducing apparent complexity.

\subsection{Software}
We used a modular software architecture when creating the system to allow for individual components to be evaluated and improved. To achieve
this modularity, we created the different components of our
software as nodes in the ROS framework \cite{ros}. As illustrated in Fig. \ref{fig:prototype}, the system consisted of several components such as attention tracking of the users, robot behavioral control, and the generation of small talk dialogue. These components collectively contributed to small talk interactions described in our initial study (Sec. \ref{sec:prototype}). In addition to these components, we introduced components of training presentation such as the session scheduler and mechanisms for generating relevant feedback to the user. 

\textbf{Delivering the Training Components.} \label{sec:training}
A scheduling node determined when the system would begin the training window. One or more training windows could be specified in the system's configurations. During the training window, the system would deliver a session, which consisted of engaging the user in several small talk conversations. At the end of each session, the system provided feedback, as detailed in Section \ref{sec:feedback}.

Recent in-home robot deployments \cite{ramnauth2022social, scassellati2018improving, georgiou2023someone} have highlighted the importance of ensuring that in-home systems are capable of discerning when it is socially appropriate to engage users. These studies have emphasized that, for effective interaction, the system must recognize contextual cues that indicate whether the user is in a setting conducive to engagement or learning. Therefore, we incorporate two classifiers: audio-based social presence classification \cite{georgiou2023someone} and person detection. The Azure Kinect captured video recordings, and image snapshots were analyzed by a pre-trained YOLO neural network \cite{farhadi2018yolov3} to estimate the number of people present. The Kinect also transcribed audio using Google's Speech-to-Text API to identify speech content. If speech was detected, the social presence classifier determined whether it represented a physically co-present conversation (e.g., a dinner party or a conversation between friends) or a media interaction (e.g., watching television or listening to the radio) \cite{georgiou2023someone}. If two or more people were detected and the audio was classified as non-media, the system assumed it was not an appropriate time to engage the user and skipped the planned interaction. However, the frequency of planned conversations increased gradually to maintain a consistent number of interactions within the training window, with intervals selected from a Gaussian distribution to avoid predictability. 

When the system was not engaging in a small talk conversation, Jibo performed an idling behavior, ranging from sleeping to looking down at the floor. When prompted to ``wake up'', Jibo looked up at the user to signal its availability to chat. The user could initiate conversation with a greeting, or Jibo would do so after some time. A linear-logarithmic growth function determined how long Jibo waited for user initiation, progressively extending this window. If the user does not respond, Jibo would prompt again. 

Following a greeting, the system randomly selected 8 to 12 rounds of conversation. It used an observer model (Sec. \ref{sec:prototype}) to ensure compliance with small talk criteria and monitor user progress. After the selected rounds are completed, Jibo transitioned to feedback mode, signaled by changing its LED ring to blue and adopting a more formal voice. 

\textbf{Generating Feedback.}\label{sec:feedback}
Feedback is essential in any training. In their study, Ramnauth et al. \cite{ramnauth2024grounded} found that quantitative definitions of small talk criteria (e.g., brevity, tone, non-specificity, coherence) effectively informed an observer model for generating feedback to the speaking model. However, these quantitative definitions do not easily translate into practical feedback for human users.

Effective user feedback should be specific, timely, constructive, objective, and respectful. We used a separate instance of GPT-3.5 that has a prompt delineating these characteristics of good micro- and macro-level feedback. \emph{Micro-level feedback} highlighted user successes after each conversation, along with one area for improvement. For instance, ``You asked great questions about my favorite hobbies. I would like to hear about your hobbies next time.''
\emph{Macro-feedback} summarized all previous conversations within a training window, providing overarching insights. For example, ``You did a great job acknowledging my interests; I enjoyed your rainy day activity suggestions. For our next conversations, try discussing your weekend plans or the weather.''
Additionally, if a small talk rule was consistently or severely broken during a training session, the robot would act out a dialogue to demonstrate following that rule. 
In these demonstrations, Jibo modulated its speech pitch and duration and changed its LED color. It even physically turned to ``face'' itself as it alternated between characters, like a puppet speaking to itself. Altogether, we aimed to explore if this generated feedback was effective, constructive, and well-received by adults with ASD. 

\textbf{Robustness for Contactless, Home Deployments}.
Robots deployed in homes require much greater robustness than those in controlled lab settings. The unstructured home environment poses challenges like power outages, fluctuating lighting, and unexpected user distractions. To enhance our system's reliability, we implemented watchdog scripts to monitor performance and remote desktop applications for troubleshooting. One script ran at the start of each training session to verify the camera, microphone, and communication with Jibo. The second script ran after each day to check the file sizes of recordings. The scripts would send an email detailing component success or failure. Remote access allowed for remote configuration; the system could be delivered to the user's home and then configured completely without in-person contact. 

\section{In-Home Deployments} \label{sec:deployment}


The study was preregistered and approved by the university's Institutional Review Board. Interested adults with ASD enrolled through the study's website promoted via channels that required a clinical diagnosis of ASD, such as ASD-specific residential facilities or employment networks. To comply with COVID-19 safety protocols, the system was designed for easy, contactless setup. It was delivered directly to participants' homes with detailed written and video instructions. Remote support was available via phone or video call, and at no point did a researcher enter participants' homes.

Participants placed the system in a comfortable room and specified a daily training window of up to three hours. The study lasted at least seven days, though adjustments were made for scheduling conflicts. After the study, participants took part in an interview to discuss their experience with the system, its effectiveness, and suggestions for improvement.


\subsection{Data Collection}
Video and audio recordings of all training interactions captured participants' responses to the robot. It is well-known that automated systems, particularly those relying on computer vision or speech recognition, often face challenges with environmental noise or the diverse behaviors exhibited by participants, especially those with ASD \cite{ramnauth2025gaze}. As a result, we opted to manually annotate the collected dataset and verify the accuracy of inputs, such as speech transcripts, before applying automated measures. Using ELAN \cite{elan}, three undergraduate research assistants annotated the start and stop times for each robot and user response. These markers allowed us to calculate users' \emph{initiation rate} defined as the proportion of conversations initiated by the user to all conversations within a session, and the duration of each \emph{conversational turn} which is the time one speaker (robot or user) spent talking before the other responded. Additionally, for each turn, a binary label indicated whether the user made eye contact with the robot. 

To ensure the reliability of these annotations, inter-rater reliability scores were calculated for each of the annotated metrics. Cohen's Kappa ($\kappa$) was used for categorical variables, such as eye contact, and the intraclass correlation coefficient (ICC) for continuous variables, such as initiation rate and conversational turn duration. The resulting $\kappa$ for eye contact was 0.95, indicating a high level of agreement between the annotators in identifying whether the user made eye contact with the robot during a given conversational turn. For the continuous metrics, the average ICC for initiation rate was 0.91, and for conversational turn duration, it was 0.93. These scores demonstrate strong consistency among annotators in evaluating the training interactions. 

Furthermore, transcripts generated by Google's Speech-to-Text were manually reviewed and corrected, particularly for users with accents or atypical intonations. Each response was assessed using the observer's evaluative metrics for small talk---brevity, tone, specificity, and coherence---as defined by Ramnauth et al. \cite{ramnauth2024grounded}. 

These annotations and observer-derived metrics enabled us explore behavioral trends over the course of the study, focusing on the frequency of user eye contact, initiative, small talk violations, and overall conversational dynamics. This quantitative analysis was further enriched by qualitative insights from post-study interviews, which explored participants' comfort with the robot, their perception of their own conversational skills, and the relevance of the robot's training and feedback.

\subsection{Participant Information}
Twenty five adults with ASD (19 males, 6 females), ranging from 18 to 68 years (\emph{M} = 32.4, \emph{SD} = 12.7) participated in the study. All had a confirmed diagnosis of ASD and completed the Autism Spectrum Quotient-10 (AQ-10) survey prior to the study. Fifteen participants were classified as high-functioning, with an average AQ-10 score of 4.8 (\emph{SD} = 1.2). The remaining ten had higher scores (\emph{M} = 7.1, \emph{SD} = 2.4) and lived with caregivers. In this group, four participants were also diagnosed with Intellectual Disability (ID), four with Attention Deficit Hyperactivity Disorder (ADHD), two with Obsessive-Compulsive Disorder (OCD), and one with Down syndrome. All with co-occurring conditions were receiving medication and specialized therapy or education at the time of this study.

\begin{figure}[t]
    \centerline{\includegraphics[width=\linewidth]{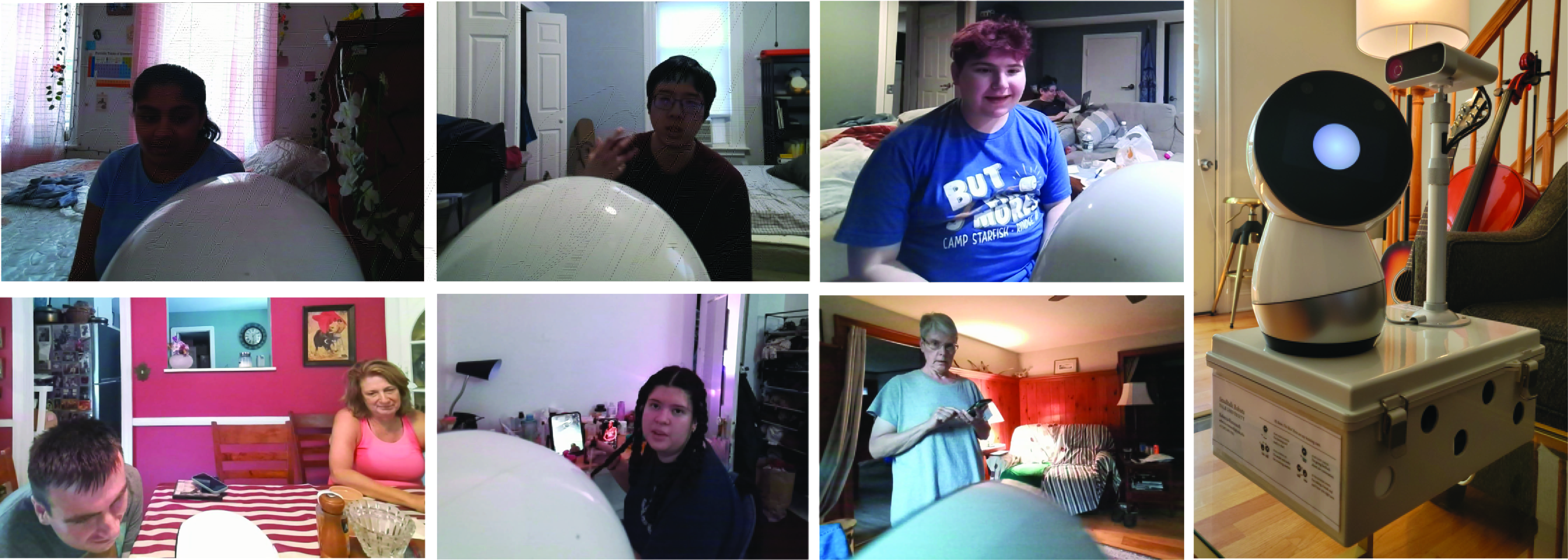}}
    \caption{\textbf{In-Home Deployments}. The collage on the left shows in-home interactions with adults with ASD from the system's point-of-view. On the right is the system placed in a user's living room after the contactless delivery.}
    \label{fig:deployment}
    \vspace{-12pt}
\end{figure}

\subsection{Results}
A total of 2,114 conversations were recorded, yielding 9,870 user responses across 225 sessions or 281.3 hours of participant interaction. Participants experienced an average of $8.9$ sessions (\emph{SD} = 2.3) over the course of the study.

\textbf{Initiating Conversations}. We analyzed the frequency of user-initiated conversations across all sessions to understand the degree of proactive engagement with the system. On average, participants initiated conversations in 34\% of interactions during their first session, rising to 64\% by the final session. Users with higher AQ-10 scores started with a lower initiation rate of 14\% in the initial session, which increased to 55\% by the end of their study (\emph{t} = 18.9, \emph{p} $\leq$ 0.0001). Conversely, high-functioning users showed an initiation rate of 50\% in the initial session, which increased to 70\% by the study's conclusion (\emph{t} = 12.4, \emph{p} $\leq$ 0.0001). All 25 participants demonstrated some improvement in their initiation rate between the initial and final sessions of the study ($\Delta$\emph{M} = 36\%, $\Delta$\emph{SD} = 20\%).

A linear regression on the initiation rate over sessions yielded a significant increase ($\beta$ = 0.07, \emph{p} $\leq$0.0001). This growth suggests that participants, irrespective of their initial engagement level, became more active in initiating small talk.

\textbf{Maintaining Eye Contact}. We assessed the effect of the training on users' ability to maintain appropriate eye contact. A linear regression analysis predicting eye contact frequency per conversational turn based on session number revealed a significant increase in participants' eye contact with the system over the course of the study ($\beta$ = 0.02, \emph{p} $\leq$ 0.001). On average, participants maintained eye contact in 24\% of turns during the first session, compared to 51\% in the final session. Additionally, all 25 participants showed improvement in eye contact frequency from the initial to final session, with an average increase of 32\% ($\Delta$\emph{SD} = 8\%).

\textbf{Sustaining Appropriate Small Talk}. We chose to separate the greeting phase from the rest of the conversation because greetings, such as ``hello'' or ``good morning,'' tend to be routine, shorter and less variable. In contrast, post-greeting responses are more varied, context-dependent, and reflect the true conversational dynamics. Greetings were identified through keyword matching and checking if the sequential response index is within the first full conversational turn. Post-greeting user responses (\emph{N} = 6,751) were assessed using the observer's metrics and behavioral annotations. 

\emph{Brevity}. The average turn duration of user responses in seconds (\emph{s}) exhibited a significant increase by session ($\beta$ = 0.77, \emph{p} $\leq$ 0.0001). The average turn duration was $4.4 s$ (\emph{SD} = 1.5) in the first session as compared to an average duration of $5.3 s$ (\emph{SD} = 2.8) in their final session. This is supplemented by the observer brevity metric based on word count; the observer flagged 2,421 user responses (21\%) as being overly verbose, and produced macro-level feedback on keeping responses more concise 61 times, or 27\% of all sessions. 

Further analysis revealed a significant increase in the robot's turn duration ($\beta$ = 0.79, \emph{p} $\leq$ 0.0001), indicating that its responses lengthened over time. Users' conversational balance, defined as the proportion of turn duration between their previous utterance and current response, also improved significantly ($\beta$ = 0.10, \emph{p} $\leq$ 0.0001). This suggests that users provided longer responses and engaged in more balanced, reciprocal interactions as the study progressed.

\emph{Non-specificity}. The observer-derived metrics for non-specificity evaluate the frequency of named entities and descriptive words in interactions. 
Due to the nature of ASD and the associated challenges with theory of mind, it is common for individuals with ASD to mention named entities (e.g., a cat named Cheddar or a street called Cedar) without considering that the robot lacks the contextual knowledge needed to respond appropriately. In the post-study interview, the mother of $P_{19}$ recounted, ``Kurt often got confused, sometimes frustrated, when the robot misunderstood him. For example, when he mentioned feeding his cat, Cheddar, the robot started asking about cheeses.'' Eight other users, all with high AQ-10 scores, demonstrated similar difficulties during their interactions. 

A linear regression on named entity counts showed a significant increase by session ($\beta$ = 0.01, $p=0.002$), as did descriptive words ($\beta$ = 0.03, \emph{p} $\leq$ 0.0001). Notably, only 16 participants exhibited these increases, with all but one categorized as high-functioning based on AQ-10 scores. The observer flagged 1,223 user responses (12\%) as overly specific, providing macro-level feedback on broadening responses in 83 sessions (37\% of total). Importantly, 85\% of these violations occurred after the first three sessions, indicating that users, particularly higher-functioning adults, felt more comfortable engaging in deeper conversations as the study progressed.


\emph{Tone}. There was no significant change in tone during the study ($\beta$ = 0.00, \emph{p} = 0.69). The observer metric for tone based on sentiment analysis yielded macro-level feedback on keeping responses more positive 27 times (12\% of all sessions). 

\emph{Coherence}. There was no significant change in user's response coherence ($\beta$ = 0.00, \emph{p} = 0.82). The observer's coherence metric based on information gain yielded feedback on keeping responses relevant 54 times (24\% of all sessions). 

\textbf{System Performance}. A key indicator of the system's performance is  sustained user engagement throughout the study. Above, we detailed trends in the frequency of user-initiated interactions and levels of eye contact with the robot. These are simultaneously considered measures of system performance and user improvement as a result of robot-assisted training.

Moreover, while Ramnauth et al. \cite{ramnauth2024grounded} showcased the effectiveness of the grounded observer in novel, in-person user interactions, we examine its performance in this context of in-home, long-term small talk training for individuals with ASD. Of all robot-delivered responses to the user ($N = 9,870$), the instances in which the observer instructed the speaking model to regenerate its proposed response ($N = 8,585$) offer valuable insights into the interaction dynamics and the effectiveness of the observer's feedback loop. In our system design, we limited the feedback loop to a maximum of three regenerations to ensure the robot delivered a timely response. From this, the first regeneration revealed the base LLM's natural tendencies, the second highlighted any gaps in the observer's feedback, and the third indicated potential issues either reintroduced or overlooked by that feedback. In our study, 65\% ($N = 6,415$) of initially proposed responses by the base model required regeneration, with brevity (45\%) and non-specificity (19\%) being the most common reasons. This indicates a tendency of the LLM to produce overly verbose and detailed responses that needed refinement. The second regeneration, which occurred in 21\% ($N = 2,072$) of these cases, primarily addressed coherence (11\%) and brevity (6\%), suggesting the observer's initial feedback in these cases resulted in less coherent, excessively terse corrections. Finally, only 1\% ($N = 98$) of proposed responses required a third regeneration, with brevity being the only recurring issue.

Since the efficacy of system prompts is often assessed through post-hoc performance or trial-and-error testing, it is challenging to quantify the exact improvement that can be achieved through a more crafted initial or feedback prompts. Nevertheless, these results demonstrate that the observer was largely successful in realigning the base LLM's deviations from appropriate small talk conventions. 

\section{Discussion}
Here, we reflect on our findings, incorporating insights gained from participant interviews. Overall, the study showed clear improvements in users' conversational skills and engagement with the training system. As training progressed, users showed greater confidence and ease in initiating conversations, maintaining eye contact, and engaging in more balanced, yet detailed conversations over time. Interestingly, while longer, more descriptive responses were flagged as violations of small talk norms---resulting in the robot giving more frequent feedback to maintain brevity---these deviations suggest an increasing preference for more substantial conversations over time. Together, these results reflect users' growing sense of ease and interest in the robot as a conversational partner.

\subsection{Post-Study Interviews}
Interviews were conducted with 24 participants with ASD\footnote{One individual was unable to participate in the post-study interview due to unforeseen personal circumstances.} and 12 primary caregivers, 10 of whom lived with a participant during the study. These semi-structured post-study interviews, conducted by a member of the research team, lasted between 30 and 45 minutes. The interviews were recorded and transcribed to ensure accuracy and participant confidentiality. Caregivers were invited to join the interview only with the prior consent of the participants. Based on the feedback collected, we performed an informal thematic analysis, categorizing the insights from participants and caregivers into three main themes, as outlined below.


\textbf{Engagement with Training}. 
Participants widely reported that the robot's structured yet dynamic interaction style helped make small talk practice approachable. Several participants mentioned that its unpredictable behavior (e.g., random sleep intervals and spontaneous prompts) mimicked real-life conversations, which kept interactions ``fresh and challenging'' ($P_{3}$). This aligns with previous findings that highlight the importance of unpredictability in training, as it more closely simulates natural, unscripted social encounters \cite{delprato2001comparisons, paul2008interventions}.

Most conversations adhered to typical small talk topics (84\% of all user responses), though some users engaged in deeper discussions, including personal experiences like bullying at work or the loss of a family member.
While some users (\emph{N} = 9) felt they had ``quickly exhausted the limited range of topics considered to be small talk'' ($P_{3}$), others (\emph{N} = 12) 
appreciated that even simple prompts about their day were ``meditative, and made me reflect on the day and my feelings in a more mindful way,'' ($P_{2}$).
Several users noted that, beyond skills training, the small talk robot provided valuable mental and emotional support. $P_{20}$ shared, ``My daily life can be isolating, so I don't often get asked how my day is going or how I'm doing [...] I enjoyed these kinds of chats when I'm making breakfast or coming home from a long day, watching TV, and there's no one else to talk to.''

The robot also served as a valuable medium for facilitating communication between caregivers and their child. Several parents (\emph{N} = 5) noted that the robot's ``neutral, non-judgmental presence'' ($P_{12}$) made it easier to initiate conversations on difficult topics. The parent of $P_{14}$ shared, ``We've had trouble talking about certain feelings, but when the robot asked about how his day was going, it opened up a way for us to talk. It felt like a safe space for him to express what he normally wouldn't share with me directly.'' 

\textbf{Perceived Skill Improvement}. Adults with ASD (\emph{N} = 18) and their caregivers (\emph{N} = 10) remarked on behavioral gains which were not captured in the observer's metrics. The robot's training was designed to simulate real conversations, complete with pauses, turns, and varied topics, which helped participants grow more confident and adept at holding discussions. For example, $P_{11}$'s parent reflected, ``My son has made great progress in understanding conversational turn-taking. He used to dominate conversations or struggle with waiting for his turn, but the training has helped him better grasp the flow of dialogue.'' Another parent observed how $P_{17}$ became more confident in her interactions with others: ``It's been incredible to see my daughter take more initiative in our family interactions. She's been practicing with Jibo, and now she actively participates in conversations at the dinner table, offering her own thoughts and even asking others about their experiences.''

\textbf{Relevance of Robot Feedback}. 
Although many robot-assisted social skills interventions for individuals with ASD exist \cite{pennisi2016autism}, most are designed primarily for children \cite{ramnauth2022social} and focus on practice rather than direct feedback. This is largely because social skills are personal and highly individualized \cite{spence2003social}, making it difficult to provide constructive feedback that resonates with each user. However, giving direct feedback is a core component of our training pedagogy. This approach to feedback could have broader implications for how we design robot-directed interventions for adult populations---while children may benefit from simpler forms of reinforcement, the robot's ability to provide explicit, personalized feedback was seen as highly valuable by many of our participants (\emph{N} = 21). 


This approach not only supported skills improvement but also encouraged cognizance of how one interacts with others. $P_{10}$ shared, ``During one session, Jibo and I discussed our favorite foods. The conversation lasted longer than usual because Jibo encouraged me to ask follow-up questions. By the end, Jibo gave me tips and even an example on how to keep conversations engaging---advice I've since used with my coworkers.'' Many users (\emph{N} = 13) commented on being more mindful and reflective about the quality of their conversations as a result of the robot's feedback. $P_{14}$ explained, ``It was new to me, practicing this kind of mindfulness—thinking more deeply about how and what to say.'' 


\subsection{Ethical Considerations}
In general, the deployment of LLM-based systems in personal settings for vulnerable users raises ethical concerns. For example, as motivation for developing the grounded observer framework, Ramnauth et al. described how inaccurate or misaligned responses could pose safety-critical risks \cite{ramnauth2024grounded}. Furthermore, the system must prioritize user privacy and security to prevent misuse of sensitive personal data. 

At the time of this study, local use of GPT-3.5 and many other LLMs were not supported. While open-source alternatives offered similar architectures, they required extensive computational resources, making them impractical for real-time interactions that rely only on the robot's on-board capabilities and a compact PC. Other stable models, including GPT-2 \cite{radford2019language} and LLAMA 2 \cite{touvron2023llama}, were trained on smaller datasets, resulting in less mature natural language capabilities and an increased risk of generating harmful language. 

To further mitigate these risks, data was transmitted to the cloud only during active participant engagement in training sessions, minimizing the amount of participant data sent throughout the in-home deployment. While our informed consent process provided participants with a clear and accessible explanation of the cloud-based LLM's use, we encourage researchers and developers to thoughtfully balance a model's readiness for real-time interaction with considerations of privacy, security, and computational feasibility.

\subsection{Study Limitations \& Directions for Future Work}
It is important to note that the present study was conceived as an exploratory investigation rather than a hypothesis-driven evaluation. Our primary objective was to demonstrate the \emph{feasibility} and process of designing, developing and deploying a social robot platform for long-term, in-home interactions with an understudied population. Still, lasting behavioral changes from training typically require months or years to manifest. Although our approximately week-long intervention captured early engagement patterns and potential novelty effects, longer-term studies are needed to assess the sustainability of any observed improvements.  

Further, given the practical challenges of establishing a comparable control condition in naturalistic settings, we adopted a within-subject design in which each participant served as their own baseline---a common approach in ASD research due to high inter-individual variability. We acknowledge the limitations of this method and refrain from making strong claims of efficacy. For instance, one may argue that increases in social initiations and eye contact could reflect familiarity or novelty effects. Interestingly, one may even expect the opposite---frequently repeated interactions, especially with a system of limited personalization or interaction memory, would result in \textit{decreased} engagement over time. While we cannot attribute user outcomes solely to the system's design, the system shows promising potential in facilitating \textit{continued} engagement with the robot-led training. Nonetheless, larger-scale studies with extended training periods will be essential for assessing the \textit{efficacy} and long-term impact of this approach.

Moreover, this study was conducted within a single country and cultural context, which may limit the generalizability of its findings to other regions and populations. Differences in communication styles, social norms, approaches to the diagnosis or treatment of ASD, and attitudes toward technology across cultures can significantly influence both the acceptance of social robots and the effectiveness of the intervention \cite{lim2021social}. To gain a more comprehensive understanding of the feasibility and efficacy of robot-assisted training, future research should replicate and expand upon this work in diverse cultural and geographic settings, accounting for the unique ways these factors may shape user engagement and outcomes.

Another limitation is that the behavioral detection and response generation of our robot system rely on models like Google’s Speech-to-Text for speech processing and OpenFace for gaze and pose estimation, which are largely trained on data from neurotypical individuals in controlled environments. Consequently, deploying such models in in-home settings with adults with ASD introduces challenges due to the unique characteristics of this population and the unpredictable nature of real-world environments \cite{ramnauth2025gaze}. For instance, some participants in our study exhibited atypical speech disfluencies, such as irregular pauses, repetition, or unexpected intonation patterns, which are common in individuals with ASD \cite{scaler2014preliminary}. These patterns can cause the system to misinterpret or entirely fail to recognize user inputs, leading to breakdowns in conversational flow. Similarly, inaccuracies in gaze estimation can result in the system failing to distinguish between the user and other visual stimuli in the environment, such as faces appearing on a TV or reflections in mirrors. This occasionally caused the robot to direct its gaze away from the user, potentially reducing the perceived quality and engagement of the interaction. Although these issues were not mentioned in our post-study interviews, we observed them while reviewing the dataset to correct errors in the generated speech transcripts. Given the growing interest in automatic behavioral annotation \cite{watson2024machine}, our observations underscore the need for model adaptations or fine-tuning to better align with the needs of diverse populations and naturalistic settings.

\section{Conclusion}
Engaging in small talk is a vital social skill that impacts everyday interactions, from making friends to job interviews. For adults with ASD, small talk poses unique challenges yet is essential for social integration and professional opportunities. In this study, we developed an autonomous, robot-assisted training platform to enhance small talk skills.  While significant, long-lasting behavioral change typically require weeks or months of training, our results showed that even within a week, adults with ASD made meaningful progress, showing increased initiative in conversations and improved eye contact.
This study makes several key contributions to our understanding of robot-assisted skills training: (1) we demonstrate how relevant training can be designed to target desired skills for an understudied population, as the majority of ASD research focuses on children and very little on adults; (2) we demonstrate the feasibility of creating a fully autonomous, plug-and-play system for daily in-home practice; (3) we emphasize the value of personalized, direct feedback beyond mere practice in social skills training; and (4) we underscore the value of an otherwise overlooked form of communication (small talk) not just for social skills development, but also for supporting positive mental and emotional outcomes for users. 


\section*{Acknowledgments}
\noindent This work was partially funded by the National Science Foundation (NSF) under grant 2106690. D. Br\v{s}\v{c}i\'{c} was supported by JST Moonshot Grant JPMJMS2011, and JSPS Kakenhi Grants JP24H00722 and JP23K11271. R. Ramnauth is supported by the NSF GRFP and the National Academies of Science (NAS) Ford Predoctoral Fellowship.

\bibliographystyle{plainnat}
\balance

\begin{thebibliography}{80}
\providecommand{\natexlab}[1]{#1}
\providecommand{\url}[1]{\texttt{#1}}
\expandafter\ifx\csname urlstyle\endcsname\relax
  \providecommand{\doi}[1]{doi: #1}\else
  \providecommand{\doi}{doi: \begingroup \urlstyle{rm}\Url}\fi

\bibitem[Akers et~al.(2016)Akers, Pyle, Higbee, Pyle, and Gerencser]{akers2016synthesis}
Jessica~S Akers, Nicole Pyle, Thomas~S Higbee, Daniel Pyle, and Kristina~R Gerencser.
\newblock A synthesis of script fading effects with individuals with autism spectrum disorder: A 20-year review.
\newblock \emph{Review Journal of Autism and Developmental Disorders}, 3:\penalty0 1--17, 2016.

\bibitem[Association et~al.(2013)]{american2013diagnostic}
American~Psychiatric Association et~al.
\newblock \emph{Diagnostic and statistical manual of mental disorders (DSM-5{\textregistered})}.
\newblock American Psychiatric Pub, 2013.

\bibitem[Babel et~al.(2021)Babel, Kraus, Miller, Kraus, Wagner, Minker, and Baumann]{babel2021small}
Franziska Babel, Johannes Kraus, Linda Miller, Matthias Kraus, Nicolas Wagner, Wolfgang Minker, and Martin Baumann.
\newblock Small talk with a robot? the impact of dialog content, talk initiative, and gaze behavior of a social robot on trust, acceptance, and proximity.
\newblock \emph{International Journal of Social Robotics}, pages 1--14, 2021.

\bibitem[Bainbridge et~al.(2008)Bainbridge, Hart, Kim, and Scassellati]{bainbridge2008effect}
Wilma~A Bainbridge, Justin Hart, Elizabeth~S Kim, and Brian Scassellati.
\newblock The effect of presence on human-robot interaction.
\newblock In \emph{RO-MAN 2008-The 17th IEEE International Symposium on Robot and Human Interactive Communication}, pages 701--706. IEEE, 2008.

\bibitem[Balderaz(2020)]{balderaz2020social}
Lindsey Balderaz.
\newblock Social skills interventions for adults with asd: a review of the literature.
\newblock \emph{Journal of Psychosocial Rehabilitation and Mental Health}, 7\penalty0 (1):\penalty0 45--54, 2020.

\bibitem[Bejerot et~al.(2014)Bejerot, Eriksson, and M{\"o}rtberg]{bejerot2014social}
Susanne Bejerot, Jonna~M Eriksson, and Ewa M{\"o}rtberg.
\newblock Social anxiety in adult autism spectrum disorder.
\newblock \emph{Psychiatry research}, 220\penalty0 (1-2):\penalty0 705--707, 2014.

\bibitem[Bellini and Peters(2008)]{bellini2008social}
Scott Bellini and Jessica~K Peters.
\newblock Social skills training for youth with autism spectrum disorders.
\newblock \emph{Child and adolescent psychiatric clinics of North America}, 17\penalty0 (4):\penalty0 857--873, 2008.

\bibitem[Bristoll and Dickinson(2015)]{bristoll2015small}
Simon Bristoll and Jules Dickinson.
\newblock Small talk, big results.
\newblock \emph{Newsli}, 92:\penalty0 6--13, 2015.

\bibitem[Burke et~al.(2010)Burke, Kraut, and Williams]{burke2010social}
Moira Burke, Robert Kraut, and Diane Williams.
\newblock Social use of computer-mediated communication by adults on the autism spectrum.
\newblock In \emph{Proceedings of the 2010 ACM conference on Computer supported cooperative work}, pages 425--434, 2010.

\bibitem[Cabibihan et~al.(2013)Cabibihan, Javed, Ang, and Aljunied]{cabibihan2013robots}
John-John Cabibihan, Hifza Javed, Marcelo Ang, and Sharifah~Mariam Aljunied.
\newblock Why robots? a survey on the roles and benefits of social robots in the therapy of children with autism.
\newblock \emph{International journal of social robotics}, 5:\penalty0 593--618, 2013.

\bibitem[Christensen-Sandfort and Whinnery(2013)]{christensen2013impact}
Robyn~J Christensen-Sandfort and Stacie~B Whinnery.
\newblock Impact of milieu teaching on communication skills of young children with autism spectrum disorder.
\newblock \emph{Topics in Early Childhood Special Education}, 32\penalty0 (4):\penalty0 211--222, 2013.

\bibitem[Coupland(2014)]{coupland2014small}
Justine Coupland.
\newblock \emph{Small talk}.
\newblock Routledge, 2014.

\bibitem[Dehqonova and Tagonova(2022)]{dehqonova2022importance}
M~Dehqonova and G~Tagonova.
\newblock The importance of didactic games in speech therapy in the development of speech in children with autism and the ability to choose effective methods.
\newblock \emph{European Journal of Innovation In Nonformal Education}, 2\penalty0 (1):\penalty0 133--137, 2022.

\bibitem[Delprato(2001)]{delprato2001comparisons}
Dennis~J Delprato.
\newblock Comparisons of discrete-trial and normalized behavioral language intervention for young children with autism.
\newblock \emph{Journal of autism and developmental disorders}, 31:\penalty0 315--325, 2001.

\bibitem[DeMatteo et~al.(2012)DeMatteo, Arter, Sworen-Parise, Fasciana, and Paulhamus]{dematteo2012social}
Francis~J DeMatteo, Patricia~S Arter, Christie Sworen-Parise, Michael Fasciana, and Marcie~A Paulhamus.
\newblock Social skills training for young adults with autism spectrum disorder: Overview and implications for practice.
\newblock \emph{National Teacher Education Journal}, 5\penalty0 (4), 2012.

\bibitem[Diehl et~al.(2012)Diehl, Schmitt, Villano, and Crowell]{diehl2012clinical}
Joshua~J Diehl, Lauren~M Schmitt, Michael Villano, and Charles~R Crowell.
\newblock The clinical use of robots for individuals with autism spectrum disorders: A critical review.
\newblock \emph{Research in autism spectrum disorders}, 6\penalty0 (1):\penalty0 249--262, 2012.

\bibitem[Duggal et~al.(2020)Duggal, Dua, Chokhani, and Sengupta]{duggal2020works}
Chetna Duggal, Bakul Dua, Ritika Chokhani, and Koyeli Sengupta.
\newblock What works and how: Adult learner perspectives on an autism intervention training program in india.
\newblock \emph{Autism}, 24\penalty0 (1):\penalty0 246--257, 2020.

\bibitem[Farhadi and Redmon(2018)]{farhadi2018yolov3}
Ali Farhadi and Joseph Redmon.
\newblock Yolov3: An incremental improvement.
\newblock In \emph{Computer Vision and Pattern Recognition}, pages 1804--02767, 2018.

\bibitem[Fiore et~al.(2013)Fiore, Wiltshire, Lobato, Jentsch, Huang, and Axelrod]{fiore2013toward}
Stephen~M Fiore, Travis~J Wiltshire, Emilio~JC Lobato, Florian~G Jentsch, Wesley~H Huang, and Benjamin Axelrod.
\newblock Toward understanding social cues and signals in human--robot interaction: effects of robot gaze and proxemic behavior.
\newblock \emph{Frontiers in psychology}, 4:\penalty0 859, 2013.

\bibitem[Garrels(2019)]{garrels2019getting}
Veerle Garrels.
\newblock Getting good at small talk: Student-directed learning of social conversation skills.
\newblock \emph{European Journal of Special Needs Education}, 34\penalty0 (3):\penalty0 393--402, 2019.

\bibitem[Georgiou et~al.(2023)Georgiou, Ramnauth, Adeniran, Lee, Selin, and Scassellati]{georgiou2023someone}
Nicholas~C Georgiou, Rebecca Ramnauth, Emmanuel Adeniran, Michael Lee, Lila Selin, and Brian Scassellati.
\newblock Is someone there or is that the tv? detecting social presence using sound.
\newblock \emph{ACM Transactions on Human-Robot Interaction}, 12\penalty0 (4):\penalty0 1--33, 2023.

\bibitem[Gerhardt and Lainer(2011)]{gerhardt2011addressing}
Peter~F Gerhardt and Ilene Lainer.
\newblock Addressing the needs of adolescents and adults with autism: A crisis on the horizon.
\newblock \emph{Journal of Contemporary Psychotherapy}, 41:\penalty0 37--45, 2011.

\bibitem[Geurts et~al.(2009)Geurts, Corbett, and Solomon]{geurts2009paradox}
Hilde~M Geurts, Blythe Corbett, and Marjorie Solomon.
\newblock The paradox of cognitive flexibility in autism.
\newblock \emph{Trends in cognitive sciences}, 13\penalty0 (2):\penalty0 74--82, 2009.

\bibitem[Graetz(2010)]{graetz2010autism}
Janet~E Graetz.
\newblock Autism grows up: Opportunities for adults with autism.
\newblock \emph{Disability \& Society}, 25\penalty0 (1):\penalty0 33--47, 2010.

\bibitem[Holmes(2005)]{holmes2005small}
Janet Holmes.
\newblock When small talk is a big deal: Sociolinguistic challenges in the workplace.
\newblock \emph{Second language needs analysis}, 344:\penalty0 371, 2005.

\bibitem[Hurlbutt and Chalmers(2004)]{hurlbutt2004employment}
Karen Hurlbutt and Lynne Chalmers.
\newblock Employment and adults with asperger syndrome.
\newblock \emph{Focus on autism and other developmental disabilities}, 19\penalty0 (4):\penalty0 215--222, 2004.

\bibitem[Ise and Iio(2021)]{ise2021social}
Naoki Ise and Takamasa Iio.
\newblock Social robot encouraging two strangers to talk with each other for their relationships.
\newblock In \emph{Companion of the 2021 ACM/IEEE International Conference on Human-Robot Interaction}, pages 144--147, 2021.

\bibitem[Jaworski(2014)]{jaworski2014silence}
Adam Jaworski.
\newblock Silence and small talk.
\newblock In \emph{Small talk}, pages 110--132. Routledge, 2014.

\bibitem[Joseph et~al.(2021)Joseph, Kearney, Brady, Downey, and Torres]{joseph2021teaching}
Brianna Joseph, Kelly~B Kearney, Michael~P Brady, Angelica Downey, and Ayse Torres.
\newblock Teaching small talk: Increasing on-topic conversational exchanges in college students with intellectual and developmental disabilities using remote audio coaching.
\newblock \emph{Behavior modification}, 45\penalty0 (2):\penalty0 251--271, 2021.

\bibitem[Kaiser et~al.(2000)Kaiser, Hancock, and Nietfeld]{kaiser2000effects}
Ann~P Kaiser, Terry~B Hancock, and Jennifer~P Nietfeld.
\newblock The effects of parent-implemented enhanced milieu teaching on the social communication of children who have autism.
\newblock \emph{Early Education and Development}, 11\penalty0 (4):\penalty0 423--446, 2000.

\bibitem[Kanero et~al.(2018)Kanero, Ge{\c{c}}kin, Oran{\c{c}}, Mamus, K{\"u}ntay, and G{\"o}ksun]{kanero2018social}
Junko Kanero, Vasfiye Ge{\c{c}}kin, Cansu Oran{\c{c}}, Ezgi Mamus, Aylin~C K{\"u}ntay, and Tilbe G{\"o}ksun.
\newblock Social robots for early language learning: Current evidence and future directions.
\newblock \emph{Child Development Perspectives}, 12\penalty0 (3):\penalty0 146--151, 2018.

\bibitem[Koegel et~al.(2014)Koegel, Park, and Koegel]{koegel2014using}
Lynn~Kern Koegel, Mi~N Park, and Robert~L Koegel.
\newblock Using self-management to improve the reciprocal social conversation of children with autism spectrum disorder.
\newblock \emph{Journal of autism and developmental disorders}, 44:\penalty0 1055--1063, 2014.

\bibitem[Kozima et~al.(2005)Kozima, Nakagawa, and Yasuda]{kozima2005interactive}
Hideki Kozima, Cocoro Nakagawa, and Yuriko Yasuda.
\newblock Interactive robots for communication-care: A case-study in autism therapy.
\newblock In \emph{ROMAN 2005. IEEE International Workshop on Robot and Human Interactive Communication, 2005.}, pages 341--346. IEEE, 2005.

\bibitem[Lab(2021)]{Jibo}
MIT~Media Lab.
\newblock Jibo social robotic research platform, 2021.
\newblock URL \url{https://www.media.mit.edu/projects/jibo-research-platform/overview/}.

\bibitem[Leyzberg et~al.(2011)Leyzberg, Avrunin, Liu, and Scassellati]{leyzberg2011robots}
Dan Leyzberg, Eleanor Avrunin, Jenny Liu, and Brian Scassellati.
\newblock Robots that express emotion elicit better human teaching.
\newblock In \emph{Proceedings of the 6th International Conference on Human-robot Interaction}, pages 347--354, 2011.

\bibitem[Lim et~al.(2021)Lim, Rooksby, and Cross]{lim2021social}
Velvetina Lim, Maki Rooksby, and Emily~S Cross.
\newblock Social robots on a global stage: establishing a role for culture during human--robot interaction.
\newblock \emph{International Journal of Social Robotics}, 13\penalty0 (6):\penalty0 1307--1333, 2021.

\bibitem[Loomes et~al.(2017)Loomes, Hull, and Mandy]{loomes2017male}
Rachel Loomes, Laura Hull, and William Polmear~Locke Mandy.
\newblock What is the male-to-female ratio in autism spectrum disorder? a systematic review and meta-analysis.
\newblock \emph{Journal of the American Academy of Child \& Adolescent Psychiatry}, 56\penalty0 (6):\penalty0 466--474, 2017.

\bibitem[Mancil(2009)]{mancil2009milieu}
G~Richmond Mancil.
\newblock Milieu therapy as a communication intervention: A review of the literature related to children with autism spectrum disorder.
\newblock \emph{Education and Training in Developmental Disabilities}, pages 105--117, 2009.

\bibitem[Matari{\'c} and Scassellati(2016)]{mataric2016socially}
Maja~J Matari{\'c} and Brian Scassellati.
\newblock Socially assistive robotics.
\newblock \emph{Springer handbook of robotics}, pages 1973--1994, 2016.

\bibitem[{Max Planck Institute for Psycholinguistics, The Language Archive, Nijmegen, The Netherlands}()]{elan}
{Max Planck Institute for Psycholinguistics, The Language Archive, Nijmegen, The Netherlands}.
\newblock Elan.
\newblock URL \url{https://archive.mpi.nl/tla/elan}.

\bibitem[McKenna et~al.(2020)McKenna, Keller, Part, Lim, Aylett, Broz, and Rajendran]{mckenna2020sorry}
Peter~E McKenna, Ingo Keller, Jose~L Part, Mei~Yii Lim, Ruth Aylett, Frank Broz, and Gnanathusharan Rajendran.
\newblock ``sorry to disturb you'' autism and robot interruptions.
\newblock In \emph{Companion of the 2020 ACM/IEEE International Conference on Human-Robot Interaction}, pages 360--362, 2020.

\bibitem[M{\"u}ller et~al.(2003)M{\"u}ller, Schuler, Burton, and Yates]{muller2003meeting}
Eve M{\"u}ller, Adriana Schuler, Barbara~A Burton, and Gregory~B Yates.
\newblock Meeting the vocational support needs of individuals with asperger syndrome and other autism spectrum disabilities.
\newblock \emph{Journal of Vocational Rehabilitation}, 18\penalty0 (3):\penalty0 163--175, 2003.

\bibitem[Nadel et~al.(2022)Nadel, Grynszpan, and Martin]{nadel2022autism}
Jacqueline Nadel, Ouriel Grynszpan, and Jean-Claude Martin.
\newblock Autism and socially interactive agents.
\newblock In \emph{The Handbook on Socially Interactive Agents: 20 years of Research on Embodied Conversational Agents, Intelligent Virtual Agents, and Social Robotics Volume 2: Interactivity, Platforms, Application}, pages 437--462. 2022.

\bibitem[Nadig et~al.(2010)Nadig, Lee, Singh, Bosshart, and Ozonoff]{nadig2010does}
Aparna Nadig, Iris Lee, Leher Singh, Kyle Bosshart, and Sally Ozonoff.
\newblock How does the topic of conversation affect verbal exchange and eye gaze? a comparison between typical development and high-functioning autism.
\newblock \emph{Neuropsychologia}, 48\penalty0 (9):\penalty0 2730--2739, 2010.

\bibitem[Nichols et~al.(2022)Nichols, Siskind, Ivanchuk, P{\'e}rez, Kamino, {\v{S}}abanovi{\'c}, and Gomez]{nichols2022hey}
Eric Nichols, Sarah~Rose Siskind, Levko Ivanchuk, Guillermo P{\'e}rez, Waki Kamino, Selma {\v{S}}abanovi{\'c}, and Randy Gomez.
\newblock Hey haru, let's be friends! using the tiers of friendship to build rapport through small talk with the tabletop robot haru.
\newblock In \emph{2022 IEEE/RSJ International Conference on Intelligent Robots and Systems (IROS)}, pages 6101--6108. IEEE, 2022.

\bibitem[Ochi et~al.(2019)Ochi, Ono, Owada, Kojima, Kuroda, Sagayama, and Yamasue]{ochi2019quantification}
Keiko Ochi, Nobutaka Ono, Keiho Owada, Masaki Kojima, Miho Kuroda, Shigeki Sagayama, and Hidenori Yamasue.
\newblock Quantification of speech and synchrony in the conversation of adults with autism spectrum disorder.
\newblock \emph{PloS one}, 14\penalty0 (12):\penalty0 e0225377, 2019.

\bibitem[Ogletree et~al.(2020)Ogletree, Price, and Campbell]{ogletree2020selective}
Billy~T. Ogletree, Johanna~R. Price, and Jonathan~M. Campbell.
\newblock A selective review of milieu interventions for adults with severe intellectual disabilities.
\newblock \emph{Current Developmental Disorders Reports}, 7:\penalty0 109--115, 2020.

\bibitem[Ohl et~al.(2017)Ohl, Grice~Sheff, Small, Nguyen, Paskor, and Zanjirian]{ohl2017predictors}
Alisha Ohl, Mira Grice~Sheff, Sarah Small, Jamie Nguyen, Kelly Paskor, and Aliza Zanjirian.
\newblock Predictors of employment status among adults with autism spectrum disorder.
\newblock \emph{Work}, 56\penalty0 (2):\penalty0 345--355, 2017.

\bibitem[Palan and Schitter(2018)]{palan2018prolific}
Stefan Palan and Christian Schitter.
\newblock Prolific. a subject pool for online experiments.
\newblock \emph{Journal of Behavioral and Experimental Finance}, 17:\penalty0 22--27, 2018.

\bibitem[Paul(2008)]{paul2008interventions}
Rhea Paul.
\newblock Interventions to improve communication in autism.
\newblock \emph{Child and adolescent psychiatric clinics of North America}, 17\penalty0 (4):\penalty0 835--856, 2008.

\bibitem[Paul et~al.(2009)Paul, Orlovski, Marcinko, and Volkmar]{paul2009conversational}
Rhea Paul, Stephanie~Miles Orlovski, Hillary~Chuba Marcinko, and Fred Volkmar.
\newblock Conversational behaviors in youth with high-functioning asd and asperger syndrome.
\newblock \emph{Journal of autism and developmental disorders}, 39:\penalty0 115--125, 2009.

\bibitem[Pennisi et~al.(2016)Pennisi, Tonacci, Tartarisco, Billeci, Ruta, Gangemi, and Pioggia]{pennisi2016autism}
Paola Pennisi, Alessandro Tonacci, Gennaro Tartarisco, Lucia Billeci, Liliana Ruta, Sebastiano Gangemi, and Giovanni Pioggia.
\newblock Autism and social robotics: A systematic review.
\newblock \emph{Autism Research}, 9\penalty0 (2):\penalty0 165--183, 2016.

\bibitem[Perkins and Berkman(2012)]{perkins2012into}
Elizabeth~A Perkins and Karen~A Berkman.
\newblock Into the unknown: Aging with autism spectrum disorders.
\newblock \emph{American journal on intellectual and developmental disabilities}, 117\penalty0 (6):\penalty0 478--496, 2012.

\bibitem[Petrolini et~al.(2023)Petrolini, Jorba, and Vicente]{petrolini2023does}
Valentina Petrolini, Marta Jorba, and Agust{\'\i}n Vicente.
\newblock What does it take to be rigid? reflections on the notion of rigidity in autism.
\newblock \emph{Frontiers in Psychiatry}, 14:\penalty0 1072362, 2023.

\bibitem[Pexman et~al.(2019)Pexman, Reggin, and Lee]{pexman2019addressing}
Penny Pexman, Lorraine Reggin, and Kate Lee.
\newblock Addressing the challenge of verbal irony: Getting serious about sarcasm training.
\newblock \emph{Languages}, 4\penalty0 (2):\penalty0 23, 2019.

\bibitem[Pullin(2010)]{pullin2010small}
Patricia Pullin.
\newblock Small talk, rapport, and international communicative competence: Lessons to learn from belf.
\newblock \emph{The Journal of Business Communication (1973)}, 47\penalty0 (4):\penalty0 455--476, 2010.

\bibitem[Radford et~al.(2019)Radford, Wu, Child, Luan, Amodei, Sutskever, et~al.]{radford2019language}
Alec Radford, Jeffrey Wu, Rewon Child, David Luan, Dario Amodei, Ilya Sutskever, et~al.
\newblock Language models are unsupervised multitask learners.
\newblock \emph{OpenAI blog}, 1\penalty0 (8):\penalty0 9, 2019.

\bibitem[Ramnauth et~al.(2022)Ramnauth, Ad{\'e}n{\'\i}ran, Adamson, Lewkowicz, Giridharan, Reiner, and Scassellati]{ramnauth2022social}
Rebecca Ramnauth, Emmanuel Ad{\'e}n{\'\i}ran, Timothy Adamson, Michal~A Lewkowicz, Rohit Giridharan, Caroline Reiner, and Brian Scassellati.
\newblock A social robot for improving interruptions tolerance and employability in adults with asd.
\newblock In \emph{2022 17th ACM/IEEE International Conference on Human-Robot Interaction (HRI)}, pages 4--13. IEEE, 2022.

\bibitem[Ramnauth et~al.(2024{\natexlab{a}})Ramnauth, Br{\v{s}}{\v{c}}i{\'c}, and Scassellati]{ramnauth2024grounded}
Rebecca Ramnauth, Dra{\v{z}}en Br{\v{s}}{\v{c}}i{\'c}, and Brian Scassellati.
\newblock A grounded observer framework for establishing guardrails for foundation models in socially sensitive domains.
\newblock \emph{arXiv preprint arXiv:2412.18639}, 2024{\natexlab{a}}.

\bibitem[Ramnauth et~al.(2024{\natexlab{b}})Ramnauth, Brscic, and Scassellati]{ramnauth_brscic_scassellati_2024}
Rebecca Ramnauth, Drazen Brscic, and Brian Scassellati.
\newblock A robot-assisted approach to small-talk training for adults with asd, Jun 2024{\natexlab{b}}.
\newblock URL \url{osf.io/kad27}.

\bibitem[Ramnauth et~al.(2025)Ramnauth, Shic, and Scassellati]{ramnauth2025gaze}
Rebecca Ramnauth, Frederick Shic, and Brian Scassellati.
\newblock Gaze behavior during a long-term, in-home, social robot intervention for children with asd.
\newblock In \emph{Proceedings of the 2025 ACM/IEEE International Conference on Human-Robot Interaction}, HRI '25, page 949–957. IEEE Press, 2025.

\bibitem[Robins et~al.(2009)Robins, Dautenhahn, and Dickerson]{robins2009isolation}
Ben Robins, Kerstin Dautenhahn, and Paul Dickerson.
\newblock From isolation to communication: a case study evaluation of robot assisted play for children with autism with a minimally expressive humanoid robot.
\newblock In \emph{2009 second international conferences on advances in computer-human interactions}, pages 205--211. IEEE, 2009.

\bibitem[Robins et~al.(2012)Robins, Dautenhahn, and Dickerson]{robins2012embodiment}
Ben Robins, Kerstin Dautenhahn, and Paul Dickerson.
\newblock Embodiment and cognitive learning--can a humanoid robot help children with autism to learn about tactile social behaviour?
\newblock In \emph{International Conference on Social Robotics}, pages 66--75. Springer, 2012.

\bibitem[Roux et~al.(2013)Roux, Shattuck, Cooper, Anderson, Wagner, and Narendorf]{roux2013postsecondary}
Anne~M Roux, Paul~T Shattuck, Benjamin~P Cooper, Kristy~A Anderson, Mary Wagner, and Sarah~C Narendorf.
\newblock Postsecondary employment experiences among young adults with an autism spectrum disorder.
\newblock \emph{Journal of the American Academy of Child \& Adolescent Psychiatry}, 52\penalty0 (9):\penalty0 931--939, 2013.

\bibitem[Scaler~Scott et~al.(2014)Scaler~Scott, Tetnowski, Flaitz, and Yaruss]{scaler2014preliminary}
Kathleen Scaler~Scott, John~A Tetnowski, James~R Flaitz, and J~Scott Yaruss.
\newblock Preliminary study of disfluency in school-aged children with autism.
\newblock \emph{International Journal of Language \& Communication Disorders}, 49\penalty0 (1):\penalty0 75--89, 2014.

\bibitem[Scassellati et~al.(2012)Scassellati, Admoni, and Matari{\'c}]{scassellati2012robots}
Brian Scassellati, Henny Admoni, and Maja Matari{\'c}.
\newblock Robots for use in autism research.
\newblock \emph{Annual review of biomedical engineering}, 14:\penalty0 275--294, 2012.

\bibitem[Scassellati et~al.(2018)Scassellati, Boccanfuso, Huang, Mademtzi, Qin, Salomons, Ventola, and Shic]{scassellati2018improving}
Brian Scassellati, Laura Boccanfuso, Chien-Ming Huang, Marilena Mademtzi, Meiying Qin, Nicole Salomons, Pamela Ventola, and Frederick Shic.
\newblock Improving social skills in children with asd using a long-term, in-home social robot.
\newblock \emph{Science Robotics}, 3\penalty0 (21), 2018.

\bibitem[Schneider and Fisk(1983)]{schneider1983attention}
Walter Schneider and Arthur~D Fisk.
\newblock Attention theory and mechanisms for skilled performance.
\newblock In \emph{Advances in psychology}, volume~12, pages 119--143. Elsevier, 1983.

\bibitem[Schwartz-Mette and Rose(2009)]{schwartz2009conversational}
Rebecca~A Schwartz-Mette and Amanda~J Rose.
\newblock Conversational self-focus in adolescent friendships: Observational assessment of an interpersonal process and relations with internalizing symptoms and friendship quality.
\newblock \emph{Journal of Social and Clinical Psychology}, 28\penalty0 (10):\penalty0 1263--1297, 2009.

\bibitem[Silver and Parsons(2022)]{silver2022perspectives}
Kate Silver and Sarah Parsons.
\newblock Perspectives of autistic adults on the strategies that help or hinder successful conversations.
\newblock \emph{Autism \& Developmental Language Impairments}, 7:\penalty0 23969415221101113, 2022.

\bibitem[Spence(2003)]{spence2003social}
Susan~H Spence.
\newblock Social skills training with children and young people: Theory, evidence and practice.
\newblock \emph{Child and adolescent mental health}, 8\penalty0 (2):\penalty0 84--96, 2003.

\bibitem[Srinivasan et~al.(2015)Srinivasan, Park, Neelly, and Bhat]{srinivasan2015comparison}
Sudha~M Srinivasan, Isabel~K Park, Linda~B Neelly, and Anjana~N Bhat.
\newblock A comparison of the effects of rhythm and robotic interventions on repetitive behaviors and affective states of children with autism spectrum disorder (asd).
\newblock \emph{Research in autism spectrum disorders}, 18:\penalty0 51--63, 2015.

\bibitem[{Stanford Artificial Intelligence Laboratory et al.}()]{ros}
{Stanford Artificial Intelligence Laboratory et al.}
\newblock Robotic operating system.
\newblock URL \url{https://www.ros.org}.

\bibitem[Touvron et~al.(2023)Touvron, Martin, Stone, Albert, Almahairi, Babaei, Bashlykov, Batra, Bhargava, Bhosale, et~al.]{touvron2023llama}
Hugo Touvron, Louis Martin, Kevin Stone, Peter Albert, Amjad Almahairi, Yasmine Babaei, Nikolay Bashlykov, Soumya Batra, Prajjwal Bhargava, Shruti Bhosale, et~al.
\newblock Llama 2: Open foundation and fine-tuned chat models.
\newblock \emph{arXiv preprint arXiv:2307.09288}, 2023.

\bibitem[van Ommeren et~al.(2012)van Ommeren, Begeer, Scheeren, and Koot]{van2012measuring}
Tineke~Backer van Ommeren, Sander Begeer, Anke~M Scheeren, and Hans~M Koot.
\newblock Measuring reciprocity in high functioning children and adolescents with autism spectrum disorders.
\newblock \emph{Journal of Autism and Developmental Disorders}, 42:\penalty0 1001--1010, 2012.

\bibitem[Watson et~al.(2024)Watson, Viana, and Zhang]{watson2024machine}
Eleanor Watson, Thiago Viana, and Shujun Zhang.
\newblock Machine learning driven developments in behavioral annotation: A recent historical review.
\newblock \emph{International Journal of Social Robotics}, pages 1--14, 2024.

\bibitem[Wei et~al.(2022)Wei, Tay, Bommasani, Raffel, Zoph, Borgeaud, Yogatama, Bosma, Zhou, Metzler, et~al.]{wei2022emergent}
Jason Wei, Yi~Tay, Rishi Bommasani, Colin Raffel, Barret Zoph, Sebastian Borgeaud, Dani Yogatama, Maarten Bosma, Denny Zhou, Donald Metzler, et~al.
\newblock Emergent abilities of large language models.
\newblock \emph{arXiv preprint arXiv:2206.07682}, 2022.

\bibitem[Wermer et~al.(2018)Wermer, Brock, and Seaman]{wermer2018efficacy}
Lauryn Wermer, Matthew~E Brock, and Rachel~L Seaman.
\newblock Efficacy of a teacher training a paraprofessional to promote communication for a student with autism and complex communication needs.
\newblock \emph{Focus on Autism and Other Developmental Disabilities}, 33\penalty0 (4):\penalty0 217--226, 2018.

\bibitem[Ying~Sng et~al.(2018)Ying~Sng, Carter, and Stephenson]{ying2018systematic}
Cheong Ying~Sng, Mark Carter, and Jennifer Stephenson.
\newblock A systematic review of the comparative pragmatic differences in conversational skills of individuals with autism.
\newblock \emph{Autism \& Developmental Language Impairments}, 3:\penalty0 2396941518803806, 2018.

\bibitem[Zimmerman(2006)]{zimmerman2006development}
Barry~J Zimmerman.
\newblock Development and adaptation of expertise: The role of self-regulatory processes and beliefs.
\newblock \emph{The Cambridge handbook of expertise and expert performance}, 186:\penalty0 705--722, 2006.

\end{thebibliography}

\end{document}